\documentclass[final]{cvpr}

\usepackage{times}
\usepackage{epsfig}
\usepackage{graphicx}
\usepackage{amsmath}
\usepackage{amssymb}
\usepackage{comment}

\usepackage[pagebackref=true,breaklinks=true,colorlinks,bookmarks=false]{hyperref}

\def\cvprPaperID{****} 
\def\confYear{CVPR 2021}

\begin{document}


\title{Interpretable COVID-19 Chest X-Ray Classification via Orthogonality Constraint \thanks{Accepted to  \href{https://www.chilconference.org/index.html}{ACM-CHIL 2021} workshop track. An extended version of this work is under consideration at Pattern Recognition Letters.}}

\author{Ella Y. Wang\\
BASIS Chandler\\
{\tt\small ellawang9@gmail.com}
\and
Anirudh Som\\
SRI International\\
{\tt\small Anirudh.Som@sri.com}
\and
Ankita Shukla\\
Arizona State University\\
{\tt\small Ankita.Shukla@asu.edu}
\and
Hongjun Choi\\
Arizona State University\\
{\tt\small hchoi71@asu.edu}
\and
Pavan Turaga\\
Arizona State University\\
{\tt\small pturaga@asu.edu}
}

\maketitle
\begin{abstract}
   Deep neural networks have increasingly been used as an auxiliary tool in healthcare applications, due to their ability to  improve performance of several diagnosis tasks. However, these methods are not widely adopted in clinical settings due to the practical limitations in the reliability, generalizability, and interpretability of deep learning based systems. As a result, methods have been developed that impose additional constraints during network training to gain more control as well as improve interpretabilty, facilitating their acceptance in healthcare community. In this work, we investigate the benefit of using Orthogonal Spheres (OS) constraint for classification of COVID-19 cases from chest X-ray images. The OS constraint can be written as a simple orthonormality term which is used in conjunction with the standard cross-entropy loss during classification network training. Previous studies have demonstrated significant benefits in applying such constraints to deep learning models. Our findings corroborate these observations, indicating that the orthonormality loss function effectively produces improved semantic localization via GradCAM visualizations, enhanced classification performance, and reduced model calibration error. Our approach achieves an improvement in accuracy of 1.6\% and 4.8\% for two- and three-class classification, respectively; similar results are found for models with data augmentation applied. In addition to these findings, our work also presents a new application of the OS regularizer in healthcare, increasing the post-hoc interpretability and performance of deep learning models for COVID-19 classification to facilitate adoption of these methods in clinical settings. We also identify the limitations of our strategy that can be explored for further research in future. 
\end{abstract}
\begin{figure}[h!]
\begin{center}
\includegraphics[width=0.99\linewidth]{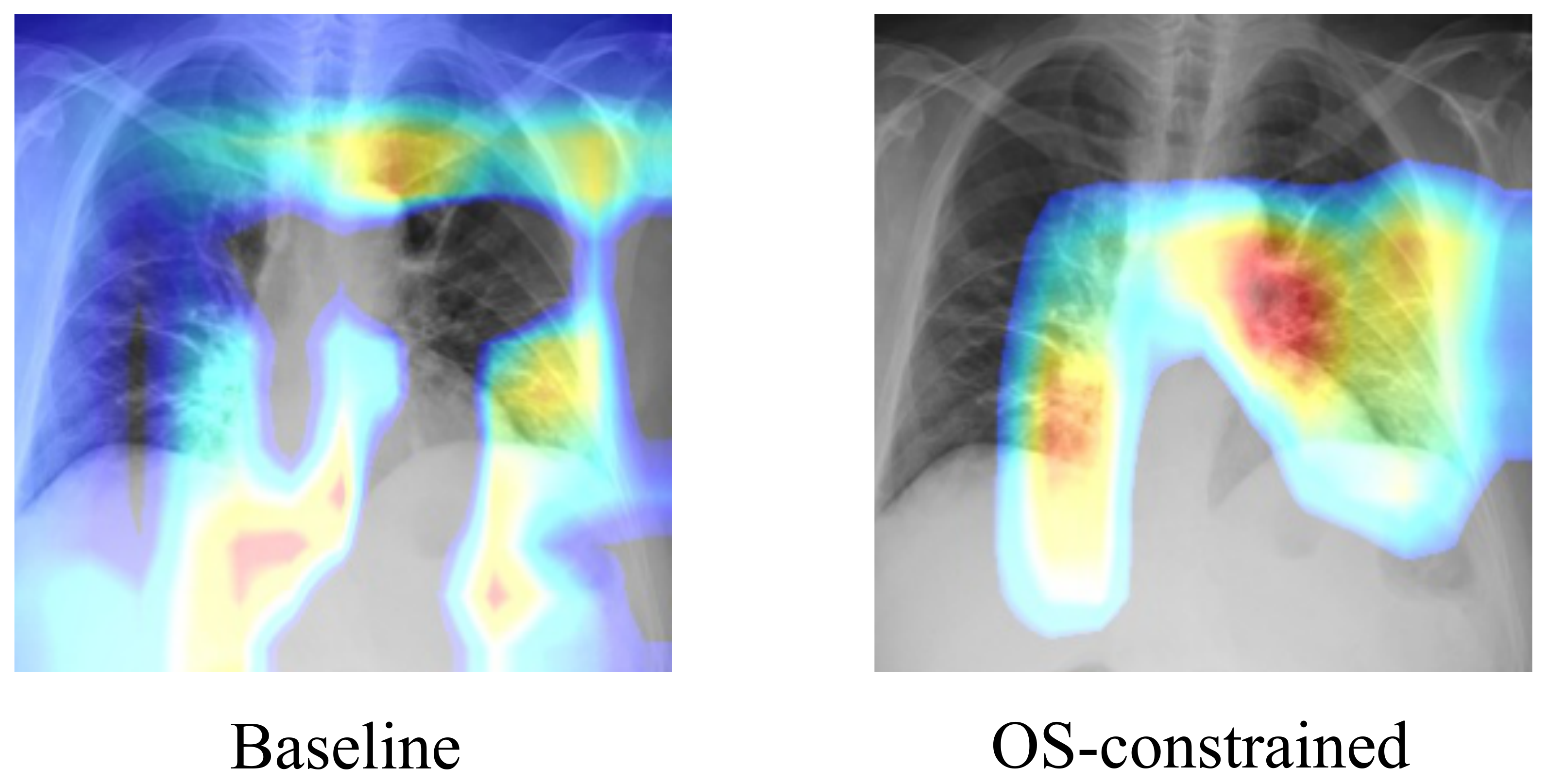}
\end{center}
	\caption{GradCAM visualization obtained from DarkCovidNet model \cite{ozturk20cnn} (serves as our baseline model) and our approach, highlighting the region of interest learned by the corresponding networks.}

    \label{preview}
\end{figure}
\section{Introduction}

Deep learning techniques have been increasingly used as an adjunct tool in medical science for developing automated solutions for disease diagnosis.
For example, they have been used to classify brain disease \cite{menikdiwela18brain}, segment lung and fundus images \cite{wu19fundus}, and detect breast cancer \cite{shen19breast}. More recently, due to the wide spread of COVID-19, deep networks have also shown to be useful in developing tools for automated detection of such cases from the chest X-ray images \cite{ozturk20cnn, wang20covidnet, minaee20deepcovid, hemdan20covidxnet}.
Thus, providing assistance in accurate and rapid diagnosis that reduces the burden on doctors as well as overcome the limitations of time consuming methods like Reverse Transcription-Polymerase Chain Reaction (RT-PCR).


COVID-19 is often diagnosed with a Reverse Transcription-Polymerase Chain Reaction (RT-PCR) using upper and lower respiratory specimens \cite{udugama20rtpcr}. However, the low sensitivity of RT-PCR (60-70\%), high false negative rates, long processing times, and shortages of testing kits hinder diagnosis and cause delays in starting treatment \cite{kanne20radiology, xie20ct}. In contrast, radiologic imaging such as computed tomography (CT) and X-ray are promising diagnostics for COVID-19. X-ray evaluations are relatively easy and fast to perform and achieve much higher sensitivity than RT-PCR, making them a more reliable and useful technology for early detection of COVID-19 \cite{ai20ctpcr}. CT is widely used in countries such as Turkey where testing kits are largely unavailable. Researchers have found that consolidation, ground-glass opacities, crazy paving pattern, and reticular pattern are common features in CT images of patients with COVID-19; Bernheim et al.\cite{bernheim20ct} observed bilateral and peripheral ground-glass opacities (GGO) as key characteristics, and Li and Xia\cite{li20ct} identified GGO and consolidation as observations. However, such subtle irregularities can only be detected by radiology experts and require valuable time, delaying diagnosis and treatment. 

Although deep learning models have achieved significant performance gains
in medical tasks, they have not been readily adopted in clinical settings due to their limited reliability, generalizability and interpretability. This limits the practical application of deep learning in healthcare due to a lack of understanding in such methods. Therefore, in order to facilitate the adoption of deep learning models
it is increasingly important to elucidate and confer trustworthiness in how these methods work.

Several deep learning approaches to automated detection of COVID-19 from chest X-ray classification have recently been developed \cite{ozturk20cnn, wang20covidnet, minaee20deepcovid, hemdan20covidxnet}. However, the post-hoc interpretability of these models is rather limited as regions of interest tend to be delocalized, resulting in less explainable interpretations of deep-classification networks in terms of semantic localization when input activation maps are visualized by the technique of Grad-CAM, which makes it difficult for radiologists to understand model decisions \cite{selvaraju17gradcam}. Furthermore, there is still much room for improvement in the overall performance and accuracy of existing models.

In this work, we aim to improve the performance of chest X-ray classification and also improve the interpretability to aid in identifying COVID cases. In the medical field, meaningful interpretability is especially important to ensure improved comprehension and explanation of model predictions for end users, such as radiologists. Interpretable deep learning models would assist healthcare personnel in driving more logical and data-driven actions, improving the quality of healthcare. 

In this work, we make use of OS parameterization to effectively train deep neural network for automated detection and classification of COVID-19 in chest X-ray images. Our work is primarily driven by the findings of earlier works by Shukla et al.\cite{shukla19os}. and Choi et al.\cite{choi20os} that have used OS constraints to improve the generalization of learned representations. Our implementation of OS constraints for chest X-ray image datasets\cite{cohen20dataset, wang17xray8} yields improvements in classification performance and better localization and preservation of regions of interest in Grad-CAM heatmap visualizations compared to baseline models. Our OS-constrained model achieved slightly higher accuracy than baseline models \cite{ozturk20cnn}, and we observe that the OS regularizer resulted in higher activation around lung areas and reduced focus on the background. These findings contribute to greater post-hoc interpretability and performance of deep learning models for detecting COVID-19. Our approach may also provide radiologists more insight into understanding classification decisions and lead to greater acceptance of deep learning models in clinical settings.

\begin{figure*}[t!]
	\begin{center}
		\includegraphics[width=1\linewidth]{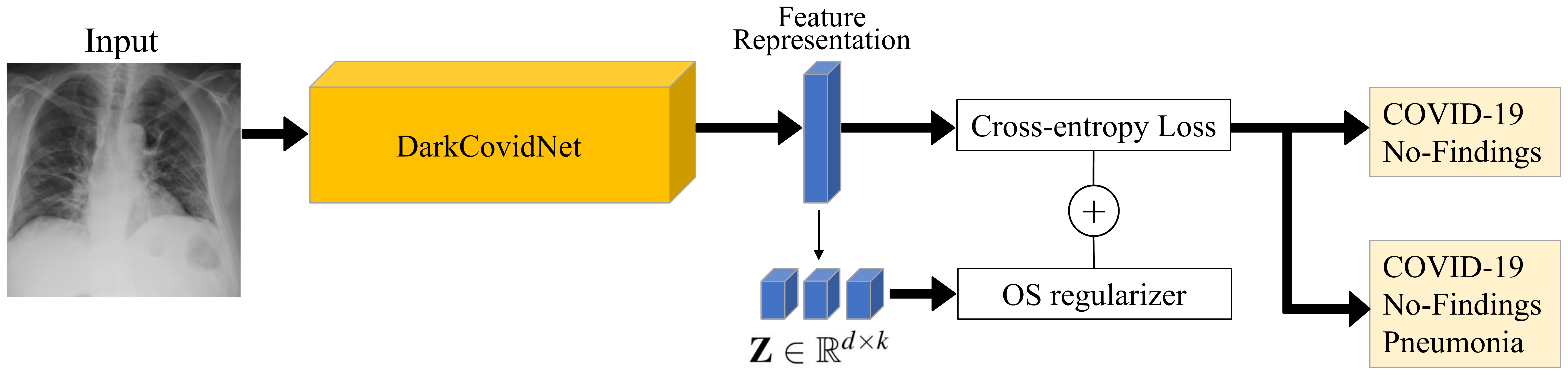}
	\end{center}
	\caption{An overview of proposed network modification on DarkCovidNet model with orthogonal spheres constraint. The $m$ dimensional  fully connected layer representation is partitioned in $k$ subsets, where each subset representation is of size $d = \frac{m}{k}$ and is stacked to represent a matrix $\mathbf{Z}$. The network is trained with cross entropy loss along with a OS regularizer that penalizes the deviation of $\mathbf{Z}$ from orthogonality condition. }
	\label{model}
\end{figure*}

\section{Related Works}
Several studies and research works have been published on the diagnosis of COVID-19 from X-ray images. Hemdan et al.\cite{hemdan20covidxnet} proposed a COVIDX-Net model made up of seven CNN models to detect COVID-19. Minaee et al.\cite{minaee20deepcovid} prepared a dataset of 5000 chest X-rays and trained four popular CNNs, reporting that ResNet18 and SqueezeNet obtained the best performance. Wang and Wong\cite{wang20covidnet} proposed a COVID-Net deep learning model to diagnose COVID-19 from X-ray images, which achieved 92.4\% accuracy in identifying healthy, non-COVID pneumonia-infected, and COVID-19-infected patients. However, these methods used limited data to develop the models. Most notably, T. Ozturk et al.\cite{ozturk20cnn} proposed the DarkCovidNet model, which has an end-to-end architecture without the need for manual feature extraction methods. Trained on a more extensive dataset of 1125 chest X-ray images\cite{cohen20dataset, wang17xray8}, the model achieved superior performance compared to other studies, obtaining 98.08\% and 87.02\% accuracy for two- and three-class classification, respectively.

In existing approaches, GradCAM\cite{selvaraju17gradcam} heat maps are used to visualize the parts of an image contributed towards the model's classification. We observe from the results of previous works that the heat maps generated from current deep learning models are highly varied. Many visualizations point to  delocalized regions of interest outside the lungs, including the shoulder bone and lung bone, despite these areas being unaffected by COVID-19. Such varied heat maps are not meaningful for post-hoc interpretation by radiologists and provide unclear insight regarding which regions of an image contributed to the final prediction. Figure \ref{preview} shows GradCAM visualizations obtained from current models, which serve as a baseline, in comparison to our orthogonal spheres approach, in which the regions highlight with red and yellow colors are considered to be important in model decisions. Baseline models convey low interpretability of model decisions, but our OS-constrained model highlights more centralized and relevant areas in the lungs.

\section{Background}

In this section, we provide a brief overview of the two components that are used in developing our strategy for chest X-ray classification for identifying COVID-19 cases.

\subsection{DarkCovidNet Model}
Ozturk et al.\cite{ozturk20cnn} proposed DarkCovidNet model that classifies chest X-ray images into three classes - no-findings, pneumonia, and COVID-19. We used DarkCovidNet model as our baseline model due to its superior performance over existing methods and modify it to incorporate the OS constraint.
The emergence of these methods is due to preference of X-ray imaging over CT scans due to their lower radiation dose. 
The DarkCovidNet model is shown to perform well with sufficient sensitivity in tasks such as detecting ground-glass opacities (GGO) in patients with COVID-19\cite{zu20ggo}. 
Further, the DarkCovidNet model was trained with a comparatively larger dataset when compared to other counterpart methods \cite{wang20covidnet, hemdan20covidxnet, narin20automatic}, developed for COVID-10 identification from Chest X-Ray images.  

Input images are of shape $256x256x3$. The DarkCovidNet model consists of 17 convolution layers and 5 pooling layers. Each DarkNet layer consists of a convolution layer, batch normalization, and a LeakyReLu operation \cite{xu15leakyrelu}. Batch normalization standardizes inputs, stabilizes the model, and reduces training time. LeakyReLU is a version of the ReLU operation \cite{agarap19relu} which has a small epsilon value to prevent dying neurons. In the DarkCovidNet model, max pooling is used in all of the pooling operations. The model ends with Flatten and Dense layers that produce the outputs. 


The last convolutional layer of the DarkCovidNet model for three classes uses $3\times3\times1$ convolutional filter with height 3, width 3, and depth 1. With this setup, the baseline DarkCovidNet model has a total of 1,170,811 parameters.  This convolutional layer is modified in our experiments to incorporate the OS constraint that requires the representation to be split into $k$ feature blocks of equal dimensions. 

\begin{table}[tb!]
	\centering
	\caption{Details of DarkCovidNet \cite{ozturk20cnn} model for 3-class classification task.} \label{model_architecture}
	\scalebox{0.90}{\begin{tabular}{ |c|c|c|c|c|c|c| } 
			\hline 
        	 Layer & Layer Type & Output Shape & Number of \\[0.5ex]
        	 Number&  &  &Trainable Parameters\\[0.5ex]
        	\hline
        	1 & Conv2D & [8, 256, 256] & 216\\[0.5ex]
        	\hline
        	2 & Conv2D & [16, 128, 128] & 1152\\[0.5ex]
        	\hline
        	3 & Conv2D & [32, 64, 64] & 4608\\[0.5ex]
        	\hline
        	4 & Conv2D & [16, 66, 66] & 512\\[0.5ex]
        	\hline
        	5 & Conv2D & [32, 66, 66] & 4608\\[0.5ex]
        	\hline
        	6 & Conv2D & [64, 33, 33] & 18,432\\[0.5ex]
        	\hline
        	7 & Conv2D & [32, 35, 35] & 2048\\[0.5ex]
        	\hline
        	8 & Conv2D & [64, 35, 35] & 18,432\\[0.5ex]
        	\hline
        	9 & Conv2D & [128, 17, 17] & 73,728\\[0.5ex]
        	\hline
        	10 & Conv2D & [64, 19, 19] & 8192\\[0.5ex]
        	\hline
        	11 & Conv2D & [129, 19, 19] & 73,728\\[0.5ex]
        	\hline
        	12 & Conv2D & [256, 9, 9] & 294,912\\[0.5ex]
        	\hline
        	13 & Conv2D & [128, 11, 11] & 32,768\\[0.5ex]
        	\hline
        	14 & Conv2D & [256, 11, 11] & 294,912\\[0.5ex]
        	\hline
        	15 & Conv2D & [128, 13, 13] & 256\\[0.5ex]
        	\hline
        	16 & Conv2D & [256, 13, 13] & 294,912\\[0.5ex]
        	\hline
        	17 & Conv2D & [3, 13, 13] & 6915\\[0.5ex]
        	\hline
        	18 & Flatten & [338] & 0\\[0.5ex]
        	\hline
        	19 & Linear & [3] & 678\\[0.5ex]
        	\hline
    \end{tabular}}
    
\end{table}

\subsection{Orthogonal Spheres}
We make use of the OS parameterization proposed by Shukla et al. \cite{shukla19os} in generative model setting and adapt it for our classification setting.  
For a given input image, let ${\mathbf{Z}}\in \mathbb{R}^{m}$ represent the output of a specific layer from the CNN model, where $m$ is the feature dimension. We partition this representation in $k$ feature blocks as $\mathbf{Z}\in\mathbb{R}^{d\times k} = [\mathbf{z}^1,\mathbf{z}^2,\dots,\mathbf{z}^k]$, where $k$ represents the number of partitions and $d$ is the dimension of each partition that is obtained as $d = \frac{m}{k}$. To make the matrix ${\mathbf{Z}}\in \mathbb{R}^{m}$ as orthogonal as possible, we regularize the off-diagonal elements in the matrix to be zero. Applying this orthogonality condition on the matrix ${\mathbf{Z}}\in \mathbb{R}^{m}$, we arrive at the simple orthonormality term shown below

\begin{align}\label{eq:OS-Loss}
L_\text{OS} &= \left\|\mathbf{Z}^{\top}\mathbf{Z}-\mathbf{I}\right\|^{2}_{F}
\end{align}

Here, $L_\text{OS}$ represents the OS regularizer and $\mathbf{I}$ represents the $k\times k$ identity matrix, with $\left\|\cdot\right\|_{F}$ being the Frobenius norm. The OS regularizer is applied along with the standard cross-entropy loss function.

This OS constraint was recently employed by Choi et al. \cite{choi20os} and have that the network  learns more diverse representations, reducing model calibration error while effectively improving the semantic localization. These improvements were shown on standard computer vision daatsets like CIFAR10\cite{krizhevsky09learning}, CIFAR100\cite{krizhevsky09learning}, SVHN\cite{netzer11reading}, and tiny ImageNet datasets\cite{deng09imagenet}. In this work, we explore and harness the capabilities of OS constraints for medical images to improve the intepretability of results, hence making them acceptable to medical practitioners.

\section{Proposed Strategy}
Deep networks are conventionally trained using the categorical-cross-entropy loss function for classification task.
However, models obtained using this loss function tend to exhibit low interpretability, feature redundancy, and poor calibration. Instead, we approach this problem with orthogonal-sphere (OS) constraints.
The OS parameterization discussed in subsection 2.3 is applied to output of the flatten layer following the last convolutional layer of the DarkCovidNet model. In doing so, we sought to reduce the number of correlated features learnt by deeper layers in the network. Our training pipeline for the proposed implementation of the OS regularizer is depicted in Figure \ref{model}.

The OS regularization function was used together with regular categorical cross-entropy loss. Thus, with $L_\text{OS}$ representing the OS regularizer, our total loss function can be characterized as 
\begin{align}\label{eq:total_loss}
L_\text{Total} &= \lambda L_\text{cross-entropy} + (1-\lambda) L_\text{OS}.
\end{align}
Here, $0 \leq \lambda \leq 1$ is a trade-off parameter. 

\section{Experimental Results}
\subsection{Dataset Description}
 Our experiments are conducted on the same dataset as used by Ozturk et al.\cite{ozturk20cnn}. The dataset has three classes: COVID-19 cases, pneumonia and healthy or no-finding. The images for COVID-19 class are obtained from an open source database of COVID-19 chest X-ray images collected by Cohen \emph{et al.} \cite{cohen20dataset}. 
 This database is continuously updated with images submitted by researchers. Currently, there are 132 X-ray images of COVID-19 diagnosis in the database,
 out of which 125 are confirmed to be positive. 
 We use these 125 images for the COVID-19 class in our experiments. In the healthy (no-findings) and pneumonia classes, 500 chest X-ray images for each class were obtained randomly from the ChestX-ray8 database collected by Wang et al. \cite{wang17xray8}, making a total of 1125 images in the dataset.




\subsection{Experimental Setup}

We performed experiments to classify COVID-19 from chest X-ray images in two different scenarios. First, we trained the DarkCovidNet model (Baseline) and OS-constrained model (Baseline + OS) to classify X-ray images into three classes: COVID-19, Pneumonia, and No-Findings. Secondly, the performance of these two models was evaluated in a classification task with two classes: COVID-19 and No-Findings. 
The performance of the models are evaluated using 5-fold cross-validation - the models are evaluated for each fold, and the average classification performance of the model is calculated. We use a 80/20 split for training and testing.

\subsection{Training Protocol and Hyper-parameter Settings}
All the experiments are conducted using a NVIDIA Tesla P100 GPU and Python 3.7 with Tensorflow 2.3.0. Our models are trained for 100 epochs using the Adam optimizer, batch-size = 32, and initial learning-rate = 0.003. We used the default Adam momentum parameters:  $\beta_{1}$ = 0.9 and $\beta_{2}$ = 0.999. Following the implementation of the DarkCovidNet model by T. Ozturk et al.\cite{ozturk20cnn}, we apply exponential learning rate decay to decay every 1000 steps with a base of 0.7. We apply batch-normalization with leaky ReLu activation with $\alpha$ = 0.1. To account for the class imbalance due to smaller number of COVID-19 images, i.e. 125 samples compared to 500 in the No-Findings and Pneumonia classes, we assign COVID-19 class four times the weight of the other two classes. The baseline DarkCovidNet model is trained using categorical cross-entropy loss function, while the OS-constrained model is trained by augmenting this loss with the orthogonality loss.
During network training, we use random horizontal flipping and slight vertical and horizontal image translation for data augmentation. When interpreting experimental results, the label ``baseline'' represents the original DarkCovidNet model trained with only cross-entropy loss; the label ``+OS'' signifies that OS-constraint applied on the baseline model, and so both cross-entropy loss and the OS regularizer are applied when training the model; the label ``+Aug'' signifies that data augmentation is used when training the model. It should be noted that DrakCovidNet model does not use data augmentation during network training.  In order to obtain flattened output with units divisible by $k$, the last layer filter in the model is modified to dimensions of $k$x$k$x1. For example, in experiments using $k$ = 4, the last convolutional layer had a filter size of 4x4x1, resulting in a flattened output size of 676 units. Experiments were performed using $k$ = 2, 3, 4, 6, 8, and 9.

\begin{figure}[t!]
\begin{center}
\includegraphics[width=1\linewidth]{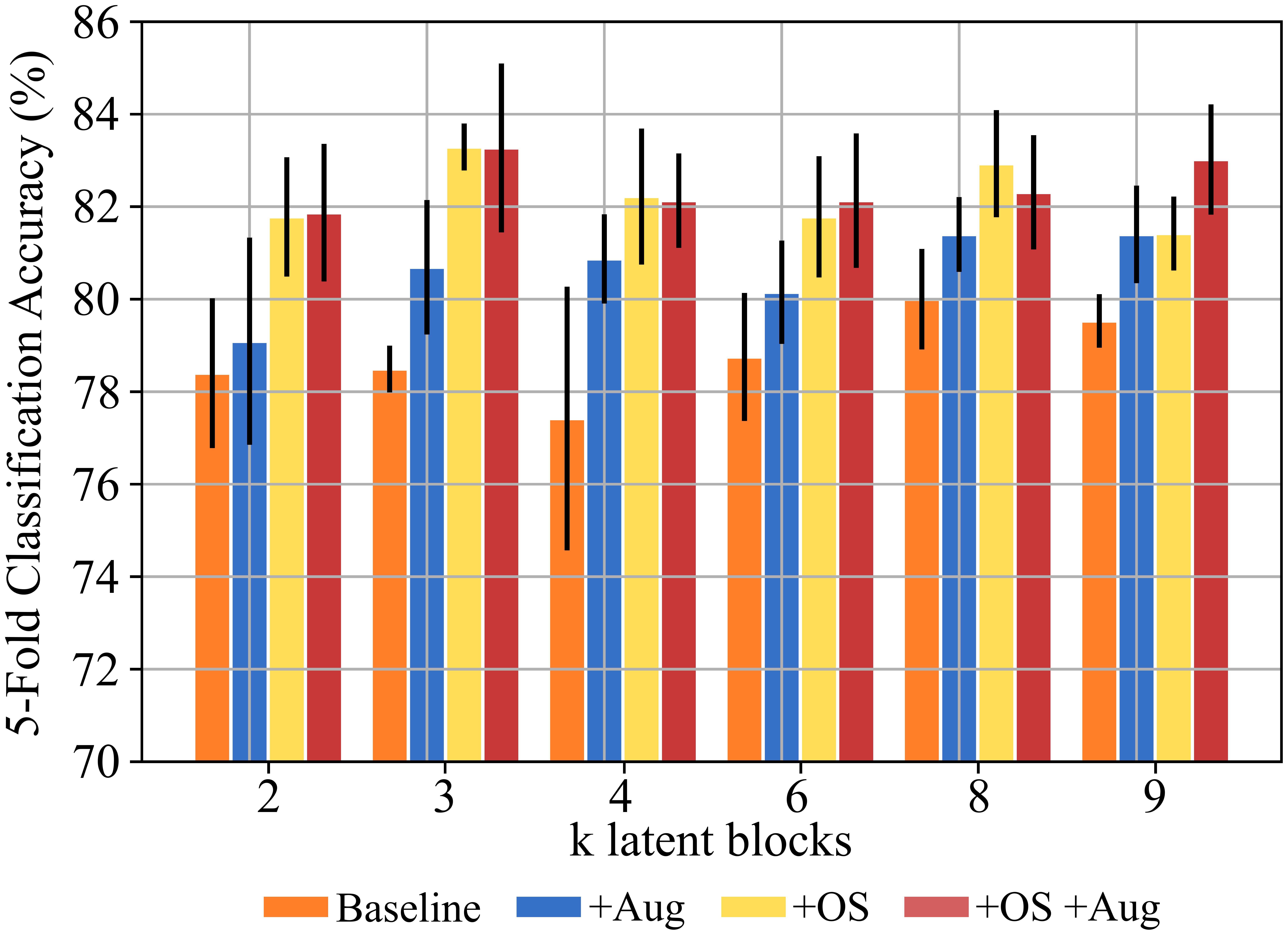}
\end{center}
	\caption{5-fold average 3-class classification accuracy for baseline and OS-constrained models trained with and without data augmentation using different values of $k$. Standard deviation bars are included.}
    \label{triple_accuracy}
    
\end{figure}
\begin{figure}[t!]
\begin{center}
\includegraphics[width=1\linewidth]{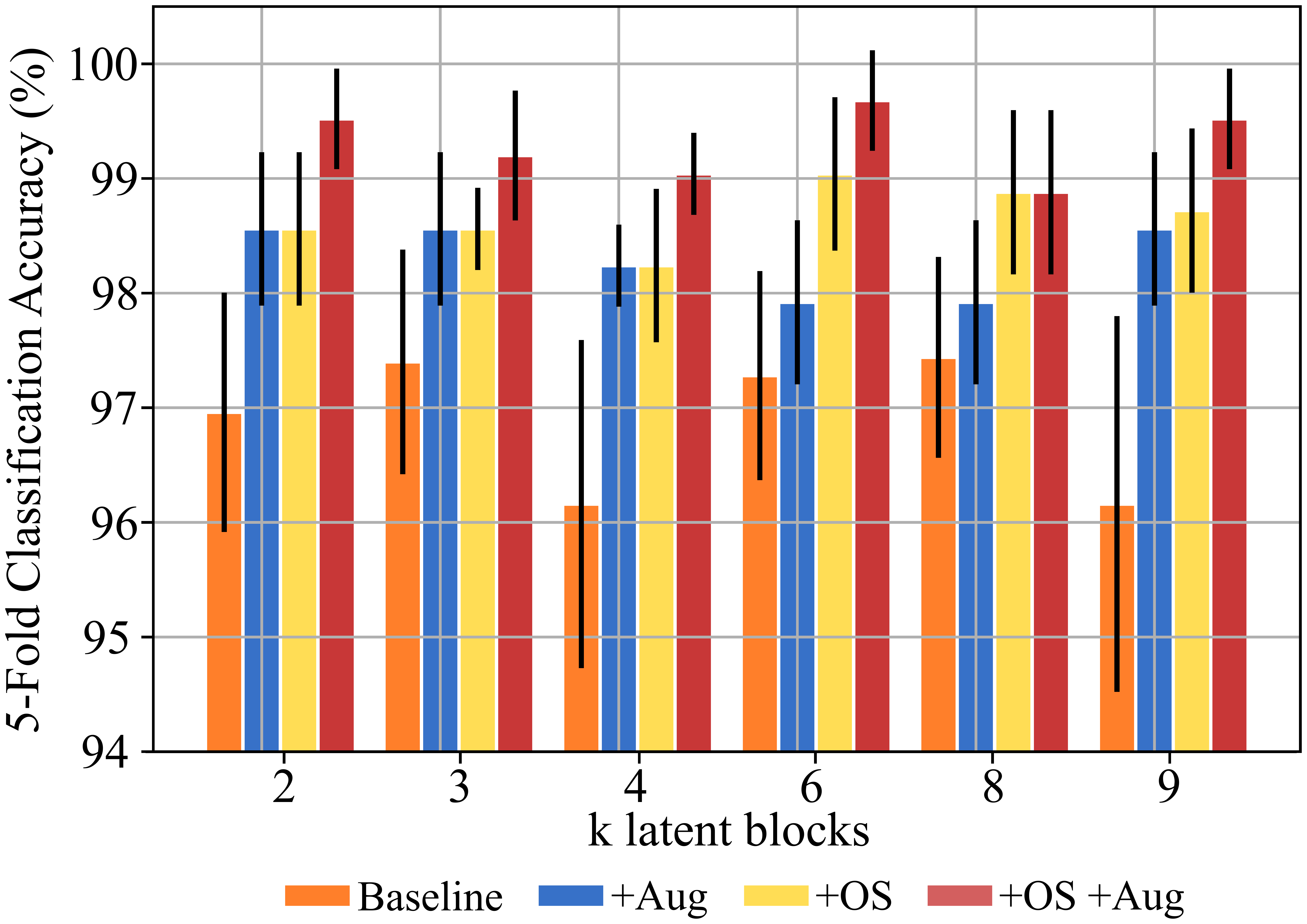}
\end{center}
	\caption{5-fold average 2-class classification accuracy for baseline and OS-constrained models trained with and without data augmentation using different values of $k$.}
    \label{binary_accuracy}
    
\end{figure}

\begin{table*}[h!]
	\centering
	\caption{Comparison of baseline and OS-constrained models with and without data augmentation in 3-class classification setting. The best results are reported in bold.} \label{triple_report} 
	\scalebox{0.99}{\begin{tabular}{ |c|c|c|c|c|c| } 
			\hline 
			& Accuracy & Precision & Recall & F1-score\\[0.5ex] 
			\hline
			Baseline & 78.49 $\pm$ 0.578 & 79.68 $\pm$ 1.31 & 79.54 $\pm$ 0.612 & 79.03 $\pm$ 0.492\\[0.5ex]
			\hline
			+Aug & 81.69 $\pm$ 1.95 & 83.21 $\pm$ 1.91 & 83.57 $\pm$ 1.95 & 83.63 $\pm$ 1.94\\[0.5ex]
			\hline
			+OS & 82.03 $\pm$ 0.798 & \textbf{83.67} $\pm$ 0.879 & 84.13 $\pm$ 0.832 & \textbf{84.82} $\pm$ 1.04\\[0.5ex]
			\hline
			+OS +Aug & \textbf{83.29} $\pm$ 1.19 & 83.14 $\pm$ 0.935 & \textbf{86.78} $\pm$ 1.03 & 84.32 $\pm$ 1.19\\[0.5ex]
			\hline
	\end{tabular}}
	
\end{table*}

\begin{table*}[h!]
	\centering
	\caption{Comparison of baseline and OS-constrained models with and without data augmentation for 2-class classification. The best results are highlighted in bold.}
	\label{binary_report}
	\scalebox{0.99}
	{\begin{tabular}{ |c|c|c|c|c|c| } 
			\hline 
			& Accuracy & Precision & Recall & F1-score\\[0.5ex] 
			\hline
			Baseline & 97.28 $\pm$ 1.62 & 97.60$\pm$ 1.41 & 89.71 $\pm$ 1.64 & 93.49 $\pm$ 1.60\\[0.5ex]
			\hline
			+Aug & 97.92$\pm$ 0.669 & 93.60 $\pm$ 0.686 & 95.90 $\pm$ 0.663 & 94.74 $\pm$ 0.677\\[0.5ex]
			\hline
			+OS & 99.04 $\pm$ 0.716 & \textbf{99.20} $\pm$ 0.711 & 96.12 $\pm$ 0.721 & 97.64 $\pm$ 0.713\\[0.5ex]
			\hline
			+OS +Aug & \textbf{99.52} $\pm$ 0.438 & 98.4$\pm$ 0.433 & \textbf{99.19} $\pm$ 0.413 & \textbf{98.8} $\pm$ 0.444\\[0.5ex]
			\hline
	\end{tabular}}
	
\end{table*}

\begin{figure}[ht!]
\begin{center}
\includegraphics[width=\linewidth]{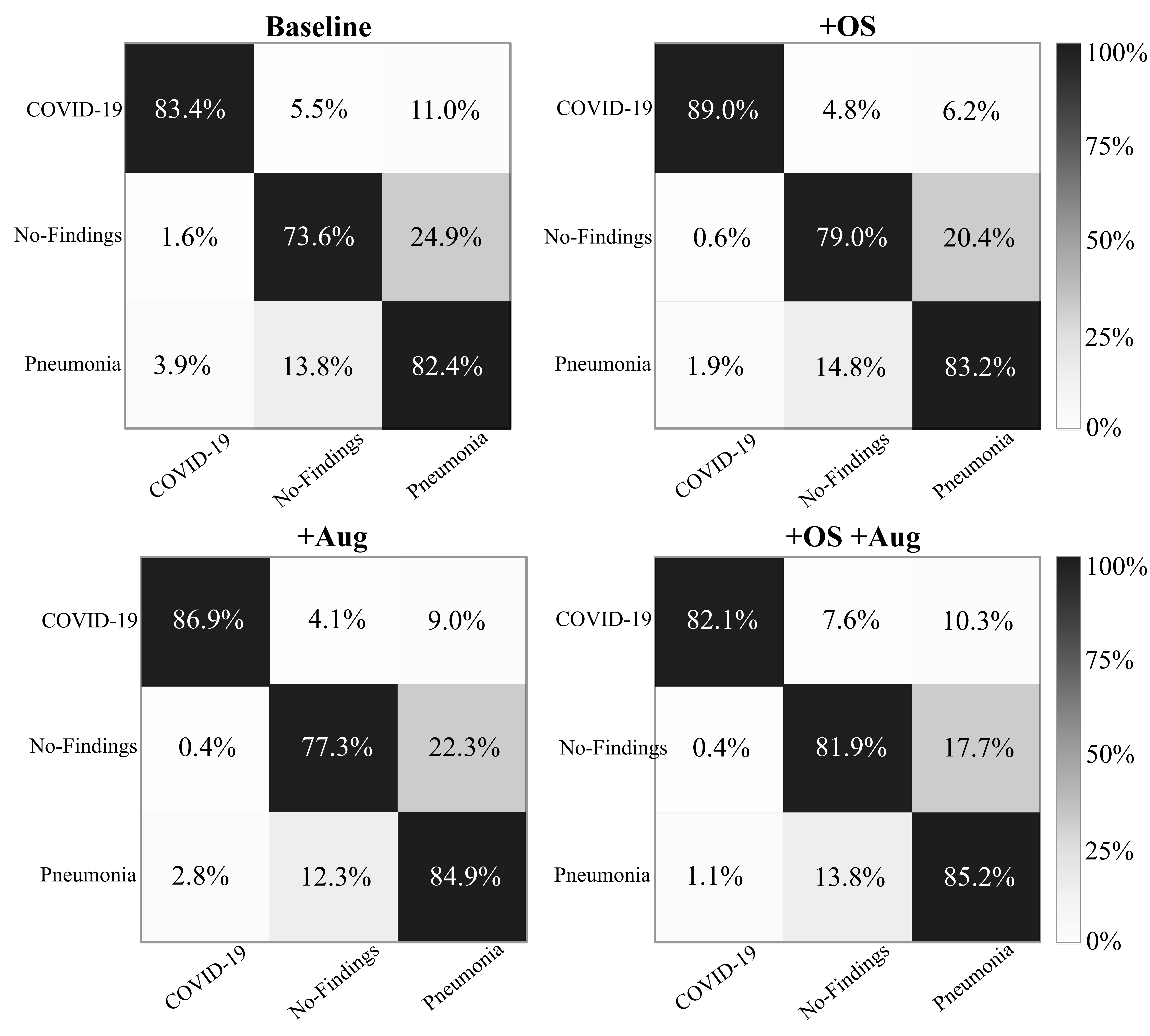}
\end{center}
	\caption{Confusion matrices for 3-class classification. The first row represents performance of regular baseline and OS-constrained models. The second row represents results obtained from the models with data augmentation applied.
	}
    \label{triple_cm}
\end{figure}

\begin{figure}[h!]
\begin{center}
\includegraphics[width=1\linewidth]{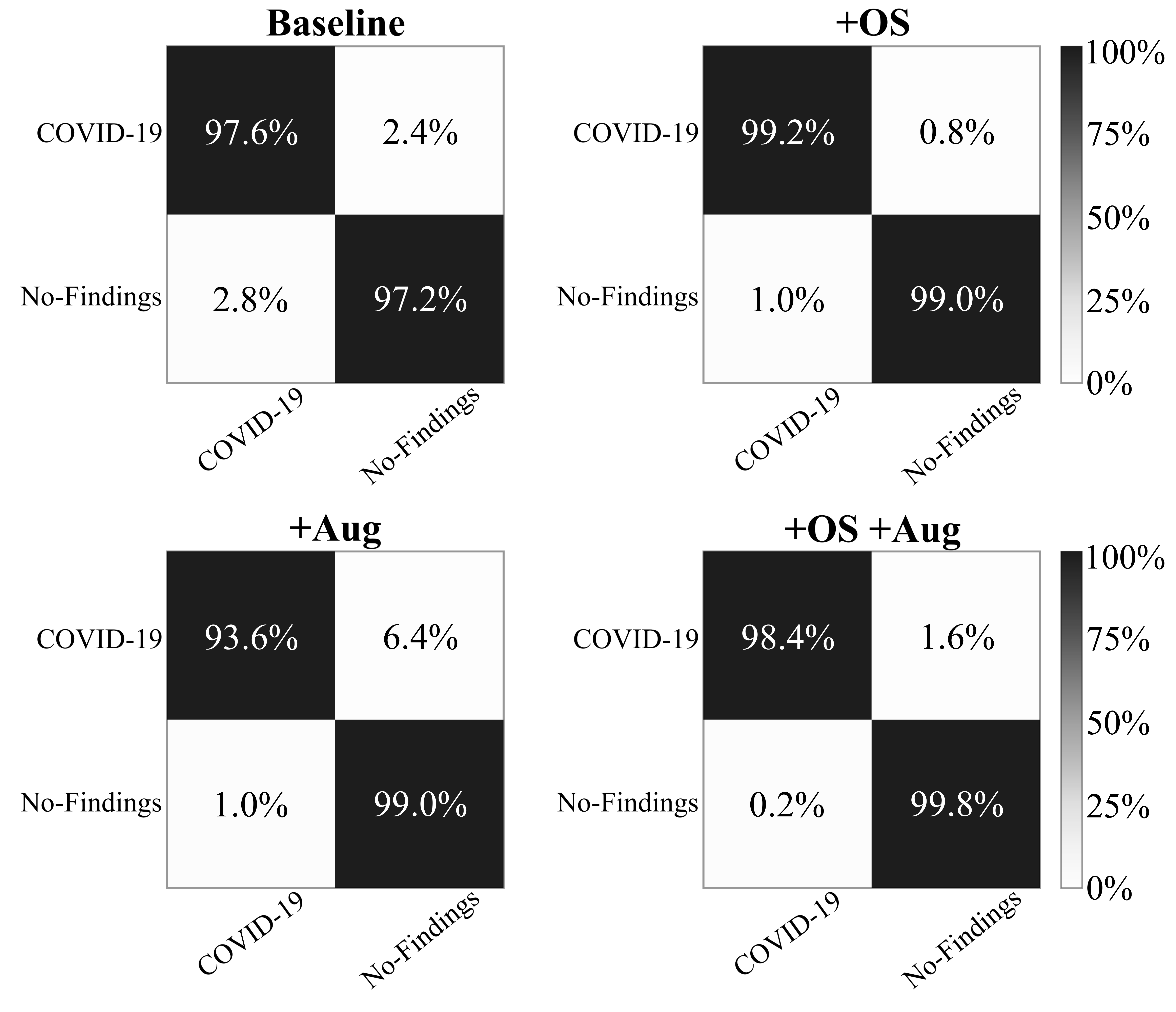}
\end{center}
	\caption{Confusion matrices for 2-class classification. The first row represents performance of baseline and OS-constrained models. The second row represents results obtained from the models with data augmentation applied.}
    \label{binary_cm}
    
\end{figure}

\begin{figure}
\begin{center}
\includegraphics[width=\linewidth]{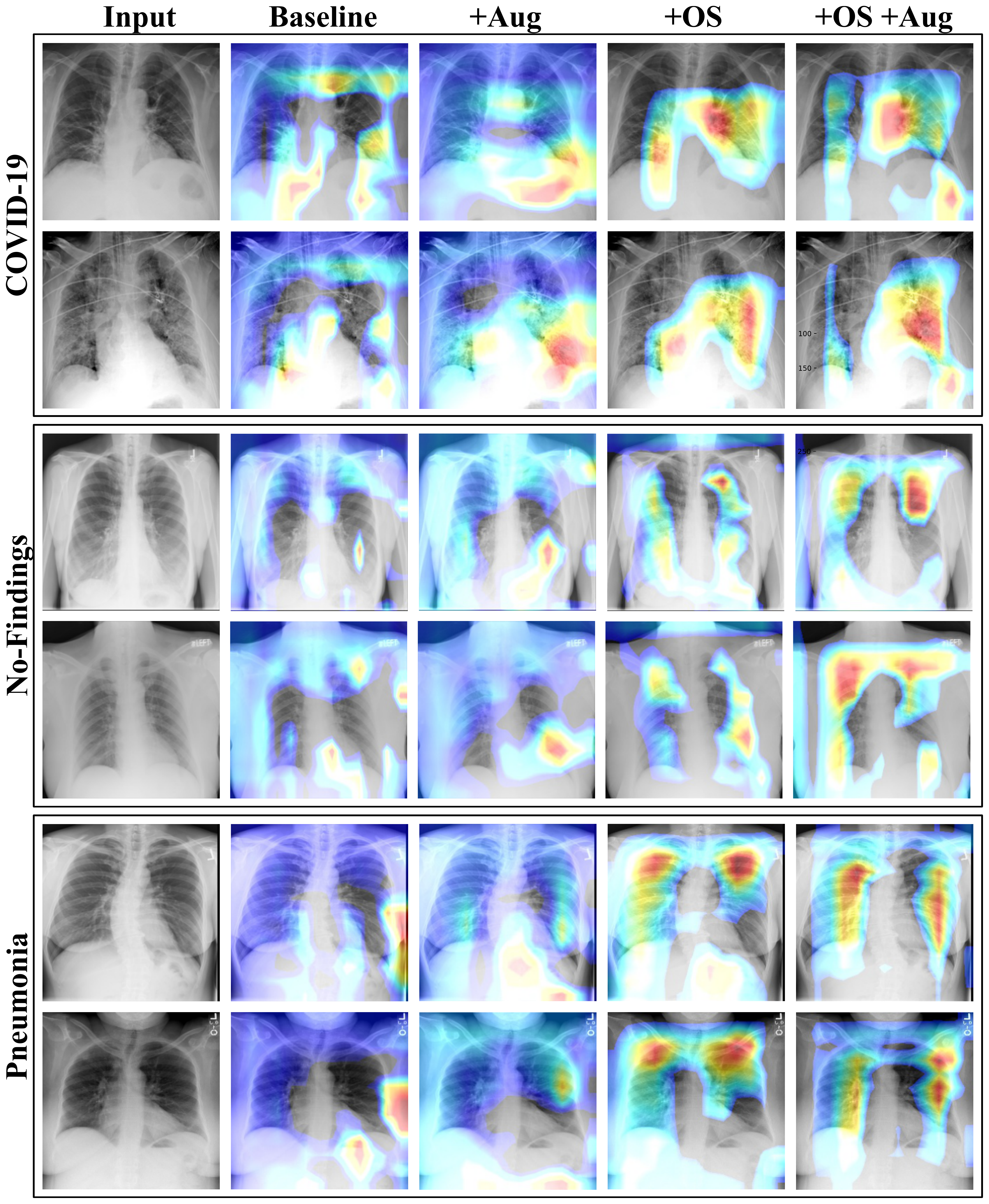}
\end{center}
	\caption{Comparison of Grad-CAM visualization obtained from DarkCovidNet (baseline) and OS-constrained models with and without data augmentation. The OS constraint uses $k$ = 3.}
    \label{gradcam_norm}
\end{figure}

\begin{figure}

\begin{center}[h!]
\includegraphics[width=\linewidth]{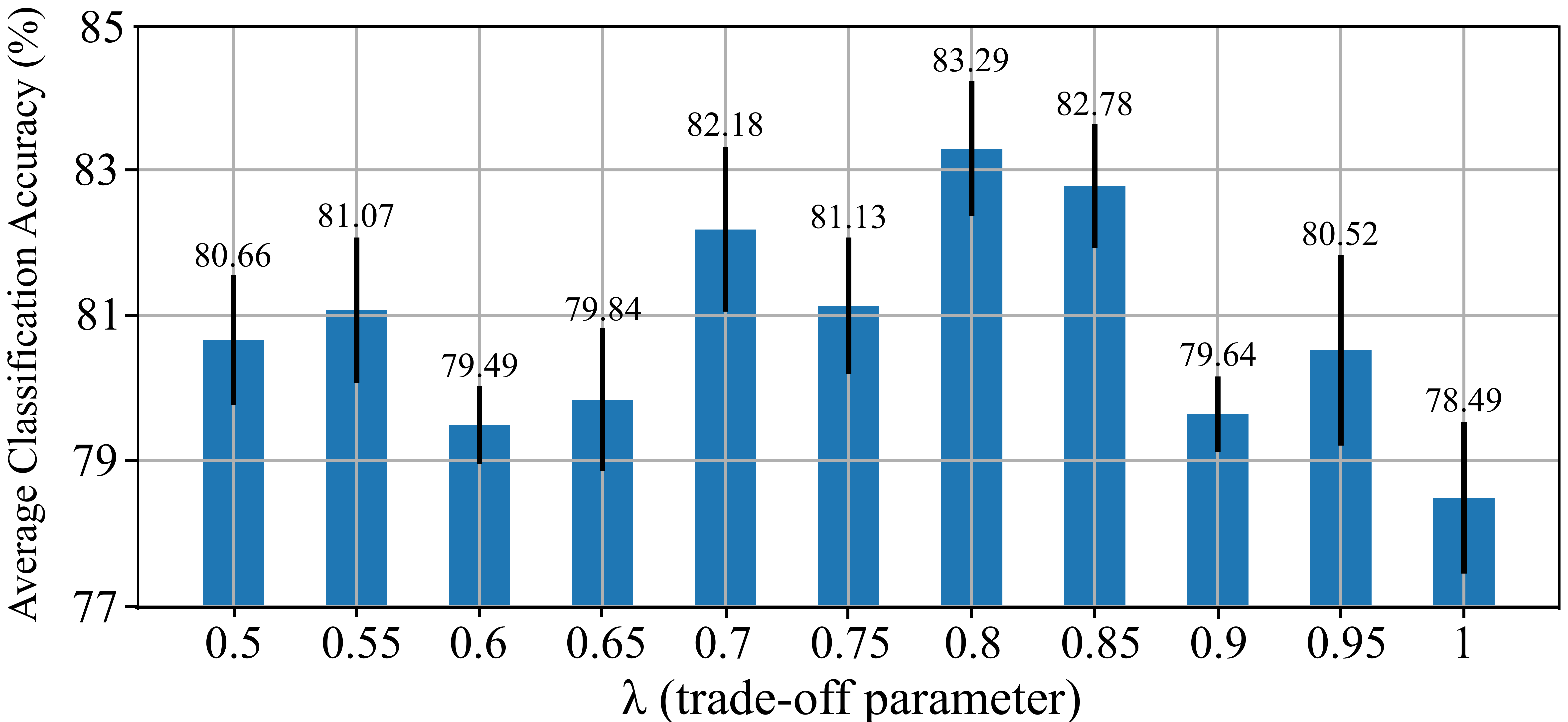}
\end{center}
	\caption{Average classification accuracy over 5 folds obtained for different values of $\lambda$. $\lambda=1$ indicates the results obtained from the baseline model trained only with cross-entropy loss. }
    \label{lambda}
\end{figure}

\begin{figure}
\begin{center}
\includegraphics[width=1\linewidth]{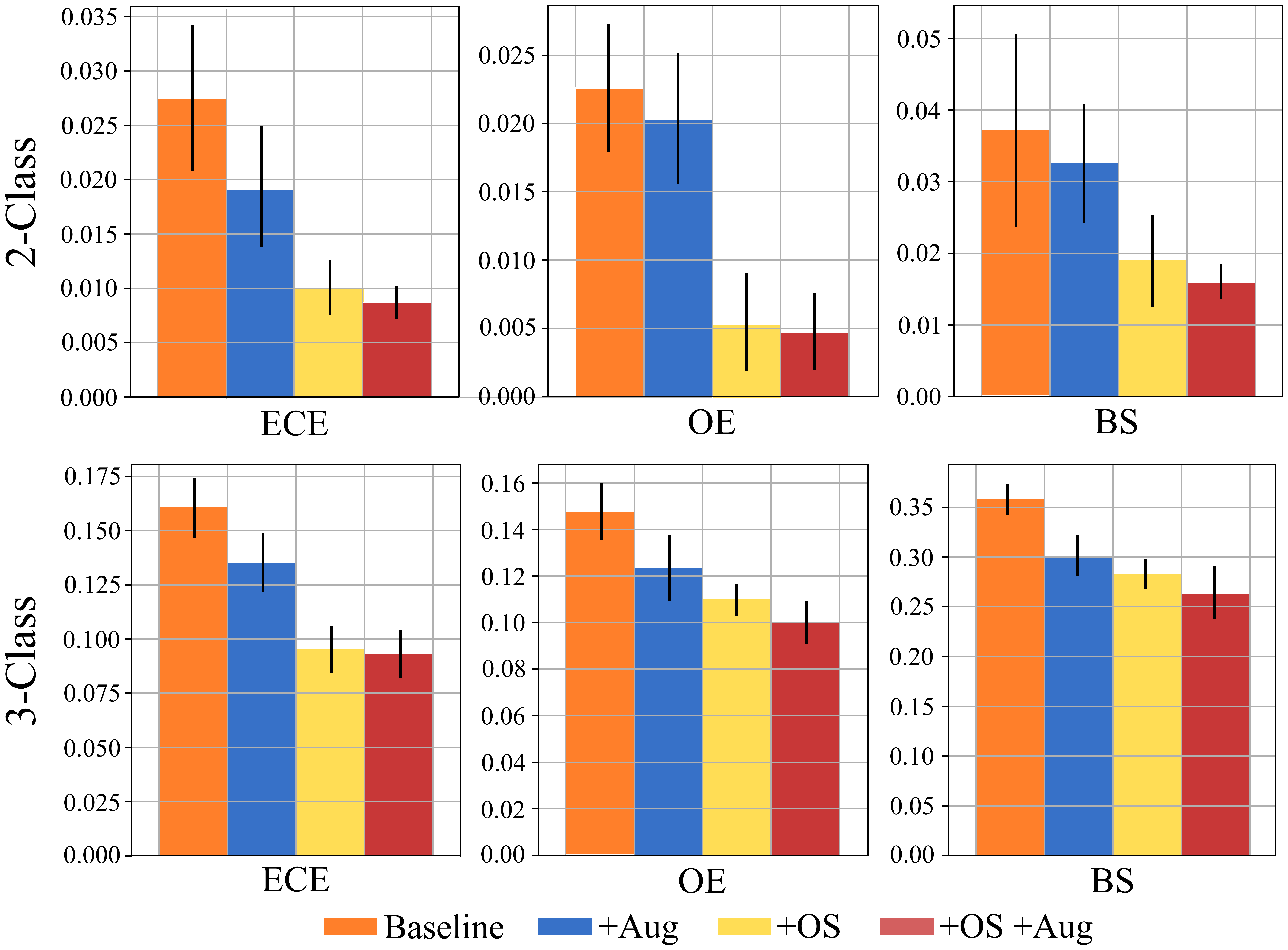}
\end{center}
	\caption{Comparison of average calibration metric scores across 5-folds for 2 (top row) and 3 (bottom row) class problem across different methods.}
    \label{calibration}
\end{figure}

\begin{figure}
\begin{center}
\includegraphics[scale= 0.1]{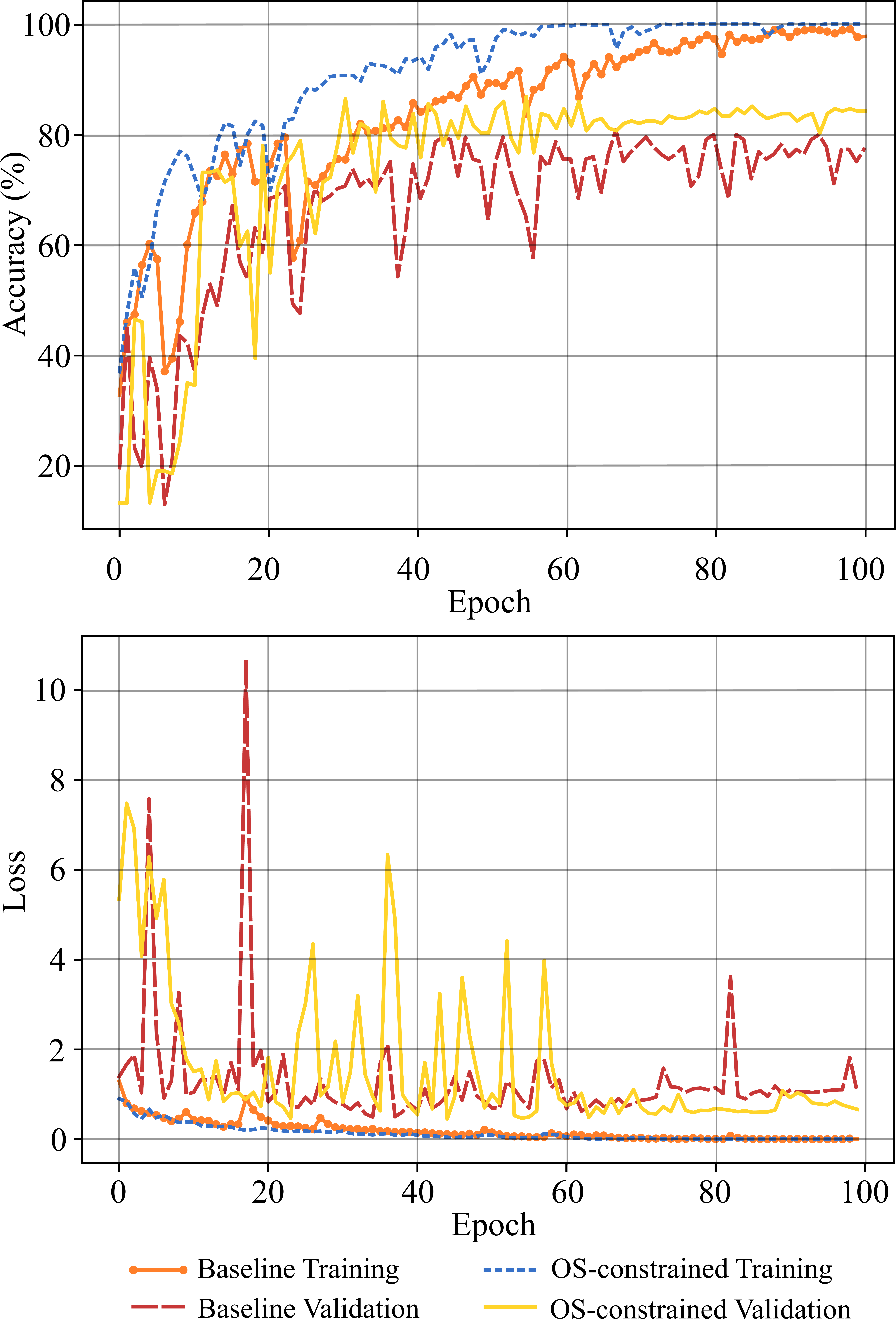}
\end{center}
	\caption{Training and validation accuracy/loss curves for the baseline and OS-constrained (baseline +OS) models.}
    \label{acc_loss}
    
\end{figure}
\begin{figure*}
\begin{center}
\includegraphics[width=1\linewidth]{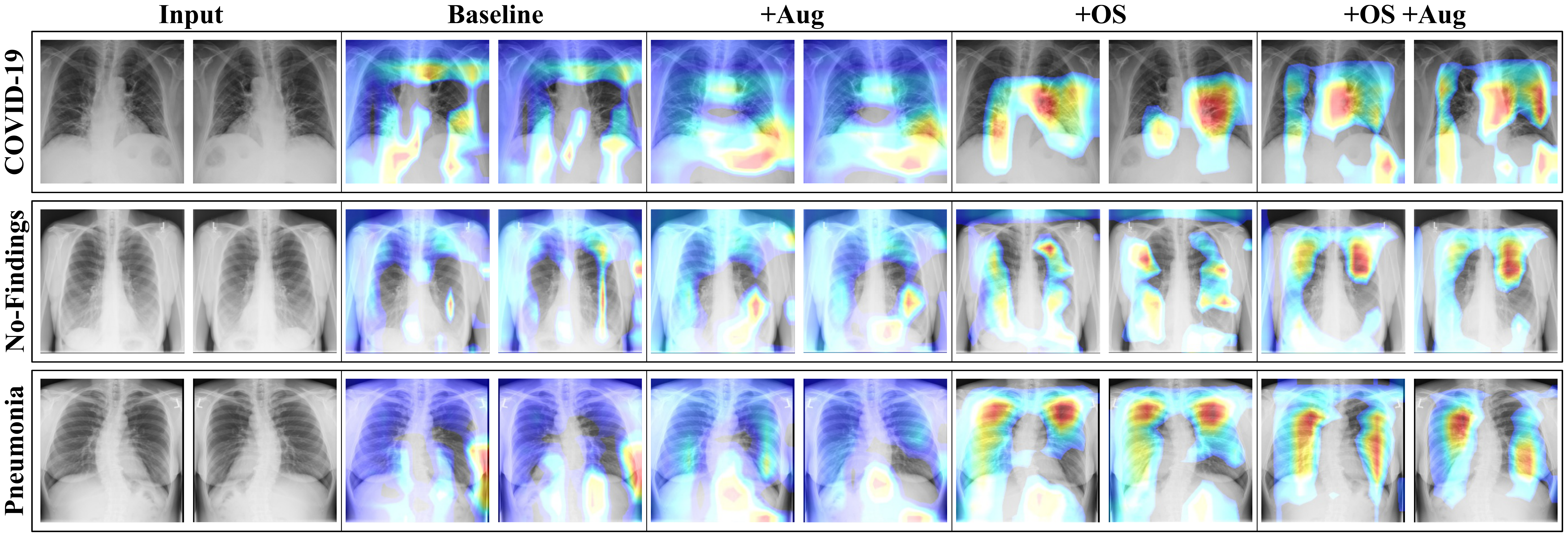}
\end{center}
	\caption{Grad-CAM visualization for baseline and OS-constrained models with and without data augmentation methods. The OS constraint uses $k$ = 3. For each pair of two columns, the first column displays the visualizations obtained from the original images, and the second column represents visualizations obtained from the horizontally flipped images.}
    \label{gradcam_flipped}
\end{figure*}

\section{Results}
\subsection{Optimizing $k$ Value}
To optimize the OS-constrained model, we performed experiments to determine the value of $k$ that results in the highest classification accuracy. As mentioned previously, $k$ represents the number of partitions of the flatten layer. The performance of the OS-constrained model with different $k$ values was evaluated for  two- and three-class classification tasks. Figure \ref{binary_accuracy} and \ref{triple_accuracy} show performance for 2 class and 3 class classification respectively. In case of 2 classes, OS-constrained model (+OS) and OS-constrained model with data augmentation (+OS+Aug), achieve marginally higher performance for $k$ = 6. On the other hand,  Figure  \ref{triple_accuracy} reports classification results of the OS-constrained model for three classes. With three classes, we find that the average accuracy is highest for $k$ = 3. These respective $k$ values are used in all other experiments. 

\subsection{Three-Class Classification}
As shown in Figure \ref{triple_accuracy}, the OS-constrained model performs slightly better than the baseline model for all values of $k$. Taking the optimal value, \emph{i.e.,} $k$ = 3, Table \ref{triple_report} shows the average accuracy, precision, recall, and F1-score across 5 folds for the OS-constrained and baseline models. The DarkCovidNet model obtains an average classification accuracy of 78.49\% and 81.69\% without and with data augmentation respectively. In comparison, the OS-constrained models obtains an average accuracy of 83.29\% and 83.27\% in the same scenario, with approximately 3-5 \% improvement over the baseline models. It can be noted that data augmentation marginally improves classification performance for both models.
Additionally we also computed the confusion matrices are shown in Figure \ref{triple_cm} for more detailed analysis of the three class problem. As pointed in \cite{ozturk20cnn}, the deep learning model is better at classifying COVID-19 than pneumonia and no-findings classes. These improvements in classification performance in the OS-constrained model can be attributed to more diverse feature representations, reduced model calibration error, and improved robustness by the OS regularizer. 
\subsection{Two-Class Classification}
Next we evaluate the performance of our OS-regularized model for the two-class classification task, involving only the COVID-19 and No-Findings classes. Figure \ref{binary_accuracy} displays the average accuracy obtained from the OS and baseline models for various $k$ values. Again, we find that the classification performance of the OS-constrained model is consistently higher than the DarkCovidNet model by a slight margin. For 2-class classification, $k$ is optimized at 6, and Table \ref{binary_report} details specific performance metrics across 5 folds. The average accuracy of the OS-constrained model was 99.04\% compared to 97.44\% by the baseline model and 99.68\% compared to 97.92\% for the models with data augmentation, reflecting a 1-2 percentage point difference. It can be noted that the performance of both OS-constrained models surpassed the 98.08\% accuracy reported by T. Ozturk et al.\cite{ozturk20cnn} for the DarkCovidNet model. 

We have also included in Figure \ref{binary_cm} the overlapped confusion matrices obtained over 5 folds, where we find that our OS-constrained model achieved slightly higher performance overall.
\subsection{Grad-CAM Visualizations}
We obtained Grad-CAM\cite{selvaraju17gradcam} heat maps to visually depict decisions made by the deep learning model. The heatmap reveals regions of the X-ray image which contributed most to the model's classification. The images in Figure \ref{gradcam_norm} represent Grad-CAM visualizations of 6 test images from the chest X-ray dataset, with 2 images per class, obtained from four experimental models for 3-class classification. Similar to the findings of T. Ozturk et al.\cite{ozturk20cnn}, the baseline DarkCovidNet model highlights more scattered areas outside the lungs, such as the chest bone, shoulders, and diaphragm, which are generally irrelevant to diagnosis and may hinder post-hoc interpretability. Although applying data augmentation to the baseline model seems to consolidate some regions, overall these areas are not helpful in understanding model decisions. Instead, the OS regularizer captures more exact and localized areas within the lobes of the lungs, suggesting improved semantic interpretation as regions of interest are better preserved. Similar to the baseline model, applying data augmentation to the OS-constrained model helped identify more relevant areas in the image. It can be noted that the OS-constrained model seems to focus more on the right side of the lung when classifying COVID-19, but emphasizes both sides of the lung for the No-Findings and Pneumonia classes. We observe that the Grad-CAM heatmaps obtained from the OS-constrained model highlight very specific lung regions which may help radiologists identify diagnostic features such as ground-glass opacities and consolidation\cite{li20ct}.

\section{Ablation Study}

\subsection{Effect of $\lambda$ parameter}
The $\lambda$ parameter in Eq. \ref{eq:total_loss} governs the contribution of OS constraint during network training. We analyzed the behaviour of network performance for different value of $\lambda$ in the range $[0,1]$
Figure \ref{lambda} shows the average accuracy obtained for different value of $\lambda$ parameter for 3-class classification performance with $k=2$. We observe that the optimal performance of the model is achieved for $\lambda$=0.8. This value of $\lambda$ used for all other experiments. 

\subsection{Model Calibration}
We also evaluate how well models were calibrated using the OS regularizer. Calibration metrics allow us to determine whether the predicted softmax scores obtained from the model are good indicators of the actual probability of the correct predictions. Our models are assessed using the Expected Calibration Error (ECE), Overconfidence Error (OE), and Brier Score (BS) \cite{guo17calibration, gneiting07strictly}. These calibration metrics is defined as:
\begin{itemize}
  \item ECE = $\sum_{m=1}^{M}\frac{|B_{m}|}{N}|\text{acc}(B_{m})-\text{conf}(B_{m})|$
  \item OE = $\sum_{m=1}^{M}\frac{|B_{m}|}{N}[\text{conf}(B_{m})\times \text{max}(\text{conf}(B_{m})-\text{acc}(B_{m}),0)]$
  \item BS = $\frac{1}{N}\sum_{n=1}^{N}\sum_{k=1}^{K}[p_{\theta}(\hat{y}_{n}=k|x_{n})-{1}(y_{n}=k)]^{2}$
\end{itemize}

Here, $B_{m}$ represents the number of predictions falling in bin $m$ and $K$ represents the number of classes. $\text{acc}(B_{m})$ denotes the accuracy of the model and $\text{conf}(B_{m})$ denotes the model's average confidence. Figure \ref{calibration} shows the calibration metric scores obtained from our baseline and OS-constrained models. Note, models with lower calibration scores are better. We find that lower calibration scores are obtained when we implement the OS regularizer with the baseline model, and data augmentation has slightly reduces calibration scores. These findings are especially significant for the 2-class classification task. 
\subsection{Effect on Network Training}
Figure \ref{acc_loss} shows the validation and training accuracy and loss curves for the baseline and OS-constrained models. The accuracy curves reveal that the OS-constrained model tends to achieve higher training and validation accuracy compared to the baseline model throughout the training period. The loss curves for both models are relatively similar, although the validation loss of the OS-constrained model shows slightly more volatility than the baseline model. 
\subsection{Horizontal Flipping}
In this subsection we study the effect of horizontally flipping input images on GRAD-Cam visualizations. Input images were mirrored across the vertical axis for testing. Using the OS-regularized and baseline models for 3-class classification task to obtain predictions, we evaluate the Grad-CAM heatmaps resulting from these modified images. Despite flipping the images, the heatmaps shown in Figure \ref{gradcam_flipped} stayed relatively consistent as those obtained from our previous experiments for all models, with highlighted regions only exhibiting slight shifts. For example, the regions emphasized by the OS-constrained models with data augmentation remained concentrated on the right side of the lung in the COVID-19 class. Since these highlighted regions were not mirrored after horizontally flipping the input images, these results suggest that despite improved performance achieved by the OS regularizer, our model still lacks robustness to transformed data. In future research, other techniques may be further explored in conjunction with OS-constraints to improve the robustness of deep learning models.
\section*{Acknowledgment}  
This work was supported in part by NSF RAPID grant 2029044.

\section{Conclusion}
In this work, we studied orthogonality constraint imposed on a deep learning model to classify COVID-19 cases from chest X-ray images. 
The proposed OS regularization yields improved performance compared to the baseline DarkCovidNet model, obtaining a classification accuracy of 83.29\% over 78.49\%  for three classes, and 99.04\% over 97.44\% accuracy for two classes without augmentation. Our OS-constrained model generates more localized and interpretable activation maps that can assist radiologists in understanding classification decisions and improving acceptance of deep learning models in the clinical settings. In future work, it is promising to explore applications of orthogonality constraints in other medical imaging tasks such as the diagnosis of chest-related diseases including pneumonia or tuberculosis.

\appendix

{\small
\bibliographystyle{ieee_fullname}
\bibliography{sample-base}

}

\end{document}